\documentclass{acm_proc_article-sp}

\newtheorem{mydef}{Definition}
\newtheorem{mythm}{Theorem}
\newtheorem{mylem}{Lemma}
\newtheorem{myobserve}{Observation}

\usepackage{amsmath,amssymb,latexsym,graphicx,url,listings, algorithm, algorithmic}
\usepackage {graphicx,ulem,textcomp}
\usepackage {todonotes}
\usepackage{hyperref,multirow}

\newcommand{\ra}{\rightarrow}

\begin{document}

\title{On The Delays In Spiking Neural P Systems \titlenote{Presented at the 6$^{th}$ Symposium on the Mathematical Aspects of Computer Science (SMACS2012), Boracay, Philippines. F.G.C. Cabarle is supported by the {DOST-ERDT program}. K.C. Bu\~ no is supported by the UP Diliman Department of Computer Science (UPD DCS). H.N. Adorna is funded by a {DOST-ERDT research grant} and the Alexan professorial chair of the {UPD DCS}. The authors also acknowledge the helpful comments of the anonymous reviewers that helped improve our work.}} %\titlenote{Partly supported by the \textit{UP Engineering Research and Development for Technology (ERDT)} program.}\\}
\numberofauthors{1} %  in this sample file, there are a *total*
\author{
% 1st. author
\alignauthor
Francis George C. Cabarle, Kelvin C. Bu\~no, Henry N. Adorna \\
Algorithms \& Complexity Lab\\
Department of Computer Science\\
University of the Philippines Diliman\\
Diliman 1101 Quezon City, Philippines\\
       \email{\{fccabarle, kcbuno, hnadorna \}@up.edu.ph }
%\and  % use '\and' if you need 'another row' of author names
}

\maketitle

%%%%%%%%%%%%%%%%%%%%%%%%%%%%%%%%%%%%%%%%%%%%%%%%%%%%%%%%%%%%%%%%%%%

\begin{abstract}
%The abstract should summarize the contents of the paper and should contain at least 70 and at most 150 words. It should be written using the \emph{abstract} environment. \keywords{We would like to encourage you to list your keywords within the abstract section}

In this work we extend and improve the results done in a previous work on simulating Spiking Neural P systems (SNP systems in short) with delays using SNP systems without delays. We simulate the former with the latter over sequential, iteration, join, and split routing. Our results provide constructions so that both systems halt at exactly the same time, start with only one spike, and produce the same number of spikes to the environment after halting.

\end{abstract}

~\\


%%%%%%%%%%%%%%%%%%%%%%%%%%%%%%%%%%%%%%%%%%%%%%%%%%%%%%%%%%%%%%%%%%%

\section{Introduction}\label{intro}

SNP systems, first presented in \cite{snp} and with some recent results in \cite{snprecent}, \cite{mem-handbook} and \cite{wsnp} (among others), are computing devices inspired by how biological neurons represent information: using electro-chemical signals called \textit{spikes}. Since spikes are indistinct, information is taken not from the spikes themselves, but from their multiplicity or time of arrival. One motivation for SNP systems (as is the case in the area of Membrane Computing \cite{introtomem} in general) is to abstract ideas from biology for computational use. For SNP systems in particular, the neuron from our brains is the motivation. It can be argued that the human brain is one (if not currently) the most complicated and powerful ``supercomputer'' known to us at the moment. The brain performs complex computations from billions of interconnected neurons while consuming only around 10 to 20 Watts of energy \cite{compspikes}, and it is small enogh to fit in our skulls. It is therefore desirable to work with as little quantity of ``energy'' as possible, and we can think of the spike in SNP systems as being such quantity.

SNP systems are Turing complete devices \cite{snp,universality} and have been used as (among others) transducers \cite{snp-ibarra}, generating vectors of numbers \cite{snp-alhazov}, as well solving hard problems \cite{snpdivision}. Spiking rules (rules that produce spikes) are usually of two types: with delays and without delays. If an SNP system has at least one rule with a delay, we refer to it as an SNP system with delay labeled as $\Pi$, otherwise it is known as an SNP system without delay labeled as $ \overline{\Pi}$.  %If a rule has a delay then it is applied after some fixed time interval, otherwise the rule is applied immediately.

In this work we extend the work presented in \cite{snp-delays}, with the goal of simulating a $\Pi$ that peforms sequential, iteration, join, and split routing with a $ \overline{\Pi}$ that performs the same routings. By \textit{routing} we mean the transfer or movement of spikes from one neuron to another. By \textit{simulation} in \cite{snp-delays} it is meant that the following two requirements are satisfied:
\begin{itemize}
\item[$ \mathbf{R_1:}$] Halting time of $\Pi$ coincides with the halting time of $ \overline{\Pi}$, or is offset either by a fixed timestep or by a function of the delays in $\Pi$, 
\item[$ \mathbf{R_2:}$] number of spikes in the final configuration of $\Pi$ is the same number in $ \overline{\Pi}$, or is offset by a function of the delays in $\Pi$.
\end{itemize}
In \cite{snp-delays}, the construction of $ \overline{\Pi}$ from $\Pi$ is such that the initial spikes of $ \overline{\Pi}$ is a function of the delay (or delays) in $\Pi$. In particular, the initial spikes of $ \overline{\Pi}$ are multiples of the delays in $\Pi$. Aside from the increased initial spike number in $ \overline{\Pi}$, the exponents in the regular expressions and the number of consumed spikes of certain spiking rules in $ \overline{\Pi}$ are also multiples of all the delays in a given routing of $\Pi$. %One open problem from \cite{snp-delays} is if we can create $ \overline{\Pi}$ with less initial spikes. Another unresolved issues in \cite{snp-delays} is the case when, given two consecutive delays $d_1$ and $d_2$ in a sequential 

%A previous work in \cite{snp-delays} presented SNP systems without delays that simulate SNP systems with delays. Assumptions were given in \cite{snp-delays} on SNP systems with delays so that SNP systems without delays simulated the delay effect for a neuron by increasing the required number of spikes to be consumed before the neuron can fire. The halting time of systems without delays with respect to systems with delays have a difference of either a constant offset or a factor of the delay. As the number of spikes required is increased, the initial number of spikes present in the system is also increased. Four possible system routings that can be simulated by the SNP system without delays were presented in \cite{snp-delays} given this solution.

%We present an alternative construction idea for simulation, with a trade-off. 
We improve the work done in \cite{snp-delays} by providing alternative constructions in this work. Our specific contributions are as follows: 
\begin{itemize}
\item we construct a $ \overline{\Pi}$ that simulates a $\Pi$ that performs sequential, iteration, join, and split routing, 
\item both $ \overline{\Pi}$ and $\Pi$ start with only one spike each in the initial neuron, 
\item halting time of $ \overline{\Pi}$ and $ \Pi$ coincide i.e. there are no offsets, 
\item number of spikes sent to the environment after halting are equal for $\Pi$ and $ \overline{\Pi}$.
\item our construction allows split routing even if the delays of the output neurons are not equal.
\end{itemize}
The trade-off is that for every delay $d$ in $\Pi$, we add $d$ neurons in $ \overline{\Pi}$. If the initial neuron of $\Pi$ has a delay, following our construction means we simply modify $\Pi$ and $ \overline{\Pi}$ such that their new halting time involves one additional time step. %, given a rule with a delay, wherein the simulation in \cite{snp-delays} needs at most one additional neuron per system routing.
%We also find that proofs for the simulations of the routings, using the idea we present in this work are, are much easier to understand...
The succeeding sections are as follows: Section \ref{prelims-sect} provides preliminaries and assumptions for our work. Section \ref{main-results-sect} presents our main results. We end with our final remarks and directions for future work in Section \ref{final-remarks-sect}.

%%%%%%%%%%%%%%%%%%%%%%%%%%%%%%%%%%%%%%%%%%%%%%%%%%%%%%%%%%%%%%%%%%%%%
\section{Preliminaries}\label{prelims-sect}

It is assumed that the readers are familiar with the basics of Membrane Computing {(a good introduction is %\cite{oxford-chap12,introtomem}
\cite{introtomem}
 with recent results and information in the P systems webpage {at {\tt http://ppage.psystems.eu/}} and a recent handbook in \cite{mem-handbook} })
and formal language theory.
We only briefly mention notions and notations which will be useful throughout the paper. 
 
Let $V$ be an alphabet, $V^*$ is the free monoid over $V$ with respect to concatenation and the identity element $\lambda$ (the empty string). The set of all non-empty strings over $V$ is denoted as $V^+$ so $V^+ = V^* - \{\lambda\}$. We call $V$ a \textit{singleton} if $V = \{a\}$ and simply write $a^*$ and $a^+$ instead of $\{a^*\}$ and $\{a^+\}$. The length of a string $w \in V^*$ is denoted by $|w|$. If $a$ is a symbol in $V$, $a^0 = \lambda$.
A language $L \subseteq V^*$ is regular if there is a regular expression $E$ over $V$ such that $L(E) = L$. A regular expression over an alphabet $V$ is constructed starting from $\lambda$ and the symbols of $V$ using the operations union, concatenation, and $+$, using parentheses when necessary to specify the order of operations. Specifically, $(i)$ $\lambda$ and each $a \in V$ are regular expressions, $(ii)$ if $E_1$ and $E_2$ are regular expressions over $V$ then $(E_1 \cup E_2)$, $E_1E_2$, and $E_1^+$ are regular expressions over $V$, and $(iii)$ nothing else is a regular expression over $V$. With each expression $E$ we associate a language $L(E)$ defined in the following way: (i) $L(\lambda) = \{\lambda\}$ and $L(a) = \{a\}$ for all $a \in V$, (ii) $L(E_1 \cup E_2) = L(E_1) \cup L(E_2)$, $L(E_1E_2) = L(E_1)L(E_2)$, and $L(E_1^+) = L(E_1)^+$, for all regular expressions $E_1$, $E_2$ over $V$. Unnecessary parentheses are omitted when writing regular expressions, and $E^+ \cup \{\lambda\}$ is written as $E^*$.  Next we have the definition for an SNP system.

\begin{mydef}[SNP system]
An SNP system of a {finite} degree $m \geq 1$  is a construct of the form
			$$\Pi=(O,\sigma_1,\ldots, \sigma_m, syn, out),$$%\todo{might consider removing $in$ from the definition just as \cite{snp-pnet} did}
%\end{definition}\todo{SNP system w/ delay?}
where:
\begin{enumerate}
\item[1.] $O=\{a\}$ is the singleton alphabet ($a$ is called \textit{spike}).

\item[2.] $\sigma_1,\ldots, \sigma_m$ are neurons of the form  $\sigma_{i}=(n_i, {R_i}),1\leq i \leq m$, where:
\begin{enumerate}
{
	\item[(a)] $n_i \geq 0$ is an integer representing the number of spikes in $\sigma_i$}
	\item[(b)] ${R_i}$ is a finite set of rules of the general form 
	
	$$E/a^c \rightarrow a^b;d$$
	
	where $E$ is a regular expression over $O$, $c \geq 1$, if $b > 0$ then $d \geq 0$ and $c \geq b$, else if $b = 0$ then $d = 0$.

\end{enumerate}

	\item[3.] $syn \subseteq \{1, 2, \ldots, m\} \times \{ 1, 2, \ldots, m \}$, $(i,i) \notin syn$ for $1 \leq i \leq m$, are synapses between neurons.

	\item[4.] $out\in \{1,2,\ldots, m\}$ is the index of the \textit{output} neuron.
\end{enumerate}

\end{mydef}

A \textit{spiking rule} is where $b \geq 1$. A \textit{forgetting rule} is a rule where $b = 0$ is written as $E/a^c \rightarrow \lambda$. If $L(E) = \{a^c\}$ then spiking and forgetting rules are simply written as $a^c \rightarrow a^b$ and $a^c \rightarrow \lambda$, respectively. Applications of rules are as follows: if neuron $\sigma_i$ contains $k$ spikes, $a^k \in L(E)$ and $k \geq c$, then the rule $E/a^c \rightarrow a^b \in R_i$ is enabled and the rule can be fired or applied. If $b \geq 1$, the application of this rule removes $c$ spikes from $\sigma_i$, so that only $k - c$ spikes remain in $\sigma_i$. The neuron sends $b$ number of spikes to every $\sigma_j$ such that $(i,j) \in syn$. The output neuron has a synapse not directed to any other neuron, only to the environment. The neuron $\sigma_1$ is referred to as the initial neuron.

If a spiking rule (forgetting rules cannot have delays) has $d = 0$, the $b$ number of spikes are sent immediately i.e. in the same time step as the application of the rule. If $d \geq 1$ and the spiking rule was applied at time $t$, then the spikes are sent at time $t+d$. From time $t$ to $t+d-1$ the neuron is said to be \textit{closed} {(inspired by the \textit{refractory period} of the neuron in biology)} and cannot receive spikes. Any spikes sent to the neuron when the neuron is closed are \textit{lost} or removed from the system. At time $t+d$ the neuron becomes \textit{open} and can then receive spikes again. The neuron can then apply another rule at time $t+d+1$.  If $b = 0$ then no spikes are produced. SNP systems assume a global clock, so the application of rules and the sending of spikes by neurons are all synchronized. 

%The \textit{nondeterminism} in SNP systems occurs when, given two rules $E_1/a^{c_1} \rightarrow a^{b_1}$ and $E_2/a^{c_2} \rightarrow a^{b_2}$, it is possible to have $L(E_1) \cap L(E_2) \neq \emptyset$. In this situation, only one rule will be nondeterministically chosen and applied. SNP systems are \textit{globally parallel} (neurons operate in parallel) but are \textit{locally sequential} (only one rule per neuron is used). In this work however we will only be considering \textit{deterministic} SNP systems.

A configuration of the system at time $k$ is denoted as $C_k = \langle n_1/t_1$, $\ldots$, $n_m/t_m, n_e \rangle$, where each element of the vector (except for $n_e$, denoting the spikes in the environment) is the configuration of a neuron $\sigma_i$, with $n_i$ spikes and is open after $t_i \geq 0$ steps. An initial configuration $C_0$ is therefore $ \langle n_1/0, \ldots, n_m/0, 0 \rangle $ since no rules whether with or without delay, have yet been applied and the environment is initially empty. %$t_i$ increases when $\sigma_i$ applies a rule with delay.
A \textit{computation} is a sequence of transitions from an initial configuration. A computation may halt (no more rules can be applied for a given configuration) or not. If an SNP system does halt, all neurons should be open.
Computation result in this work is obtained by checking the number of spikes in the environment once the system halts.

\begin{figure}[h]
\unitlength=1mm
\special{em:linewidth 0.4pt}
\linethickness{0.4pt}
\begin{picture}(116.33,12)(1,0)

%ovals i.e. neurons
\put(10.83,5.17){\oval(16.00,8.67)[]} %\sigma_1
\put(34.83,5.17){\oval(20.00,8.67)[]} %\sigma_2
\put(58.83,5.17){\oval(16.00,8.67)[]} %\sigma_3

%arcs i.e. synapses
\put(18.67,4.67){\vector(1,0){6.33}} %\sigma_1 -> \sigma_2
\put(44.67,4.67){\vector(1,0){6.33}} %2 -> 3
\put(66.67,4.67){\vector(1,0){6.33}} %3 -> env

%contents of neurons
%\sigma_1
\put(2,2.33){\makebox(0,0)[cc]{$1$}}
\put(10.33,8.67){\makebox(0,0)[cc]{$a$}}
\put(10.33,5.67){\makebox(0,0)[cc]{$a^+/a\ra a$}}
%\sigma_2
\put(24,2.33){\makebox(0,0)[cc]{$2$}}
\put(34.33,5.67){\makebox(0,0)[cc]{$a^+/a\ra a;2$}}
%\sigma_3
\put(50,2.33){\makebox(0,0)[cc]{$3$}}
\put(58.33,5.67){\makebox(0,0)[cc]{$a^+/a\ra a$}}
\end{picture}
\caption{SNP system with delay $\Pi_0$. }\label{example_snp-fig}
\end{figure}
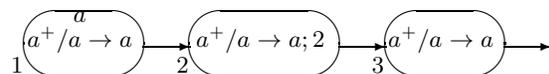

As an example, let us have an SNP system shown in Figure \ref{example_snp-fig} formally defined as follows: $\Pi_0$ = $(O, \sigma_1, \sigma_2, \sigma_3, syn, out )$ where $\sigma_1 = ( 1, a^+/a \rightarrow a) $, $ \sigma_2 = ( 0, a^+/a \rightarrow a;2)$, $ \sigma_3 = ( 0, a^+/a \rightarrow a )$,  $ syn = \{ (1,2)$, $(2,3) \}$, initial neuron is $\sigma_1$, and $out = 3$. Only neuron $\sigma_1$ has one spike initially and only $\sigma_2$ has a rule with a delay $d = 2$. We have $C_0 = \langle 1/0, 0/0, 0/0, 0 \rangle $. At the next step, $\sigma_1$ can use its rule (it has at least one spike) and consumes one spike and sends one spike immediately to $\sigma_2$ so we have $C_1 = \langle 0/0, 1/0, 0/0, 0 \rangle$. At step 2, $\sigma_2$ consumes its spike and closes for 2 time steps, so $C_2 = \langle 0/0, 0/2, 0/0, 0 \rangle $. At step 3 we have $C_3 = \langle$ $0/0, 0/1,$ $0/0, 0 \rangle $. At time step 4, $\sigma_2$ opens and sends one spike to $\sigma_3$, so $C_3 = \langle 0/0, 0/0, 1/0, 0 \rangle $. Finally, at time step 5 the output neuron sends one spike to the environment, $\Pi_0$ halts and we have $C_5 = \langle 0/0, 0/0, 0/0, 1 \rangle $.

%This WORKS!...
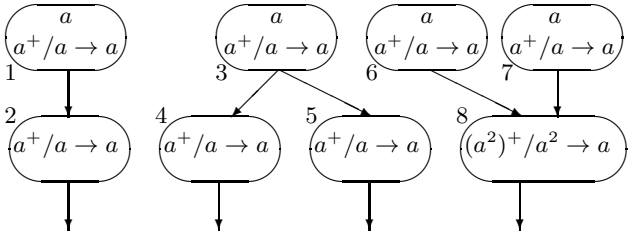
\begin{figure}[h]
\unitlength=1mm
\special{em:linewidth 0.4pt}
\linethickness{0.4pt}
\begin{picture}(116.33,35)(1,0)

%ovals i.e. neurons
\put(10.83,12.17){\oval(16.00,8.67)[]} %\sigma_2
\put(10.33,27.00){\oval(16.00,8.67)[]} %\sigma_1

\put(38.33,27.00){\oval(16.00,8.67)[]} %\sigma_3
\put(30.83,12.17){\oval(16.00,8.67)[]} %\sigma_4
\put(50.83,12.17){\oval(16.00,8.67)[]} %\sigma_5

\put(58.33,27.00){\oval(16.00,8.67)[]} %\sigma_6
\put(76.33,27.00){\oval(16.00,8.67)[]} %\sigma_7
\put(73.83,12.17){\oval(22.00,8.67)[]} %\sigma_8

%arcs i.e. synapses
\put(10.67,22.67){\vector(0,-1){6.33}} %\sigma_1 -> \sigma_2
\put(10.67,7.67){\vector(0,-1){6.33}} %\sigma_2 -> env

\put(38.67,22.67){\vector(-1,-1){6.33}} %3 -> 4
\put(38.67,22.67){\vector(2,-1){12.33}} %3 -> 5
\put(30.67,7.67){\vector(0,-1){6.33}} %4 -> env
\put(50.67,7.67){\vector(0,-1){6.33}} %5-> env

\put(58.67,22.67){\vector(2,-1){12.33}} %6 -> 8
\put(75.67,22.67){\vector(0,-1){6.33}} %7 -> 8
\put(70.67,7.67){\vector(0,-1){6.33}} %8-> env

%contents of neurons
%\sigma_1
\put(3,22.33){\makebox(0,0)[cc]{$1$}}
\put(10.33,29.67){\makebox(0,0)[cc]{$a$}}
\put(10.33,25.67){\makebox(0,0)[cc]{$a^+/a\ra a$}}
%\sigma_2
\put(3.00,16.67){\makebox(0,0)[cc]{$2$}}
\put(10.33,12.67){\makebox(0,0)[cc]{$a^+/a\ra a$}}
%\sigma_3
\put(31,22.33){\makebox(0,0)[cc]{$3$}}
\put(38.33,29.67){\makebox(0,0)[cc]{$a$}}
\put(38.33,25.67){\makebox(0,0)[cc]{$a^+/a\ra a$}}
%\sigma_4
\put(23.00,16.67){\makebox(0,0)[cc]{$4$}}
\put(30.33,12.67){\makebox(0,0)[cc]{$a^+/a\ra a$}}
%\sigma_5
\put(43.00,16.67){\makebox(0,0)[cc]{$5$}}
\put(50.33,12.67){\makebox(0,0)[cc]{$a^+/a\ra a$}}
%\sigma_6
\put(51,22.33){\makebox(0,0)[cc]{$6$}}
\put(58.33,29.67){\makebox(0,0)[cc]{$a$}}
\put(58.33,25.67){\makebox(0,0)[cc]{$a^+/a\ra a$}}
%\sigma_7
\put(69,22.33){\makebox(0,0)[cc]{$7$}}
\put(75.33,29.67){\makebox(0,0)[cc]{$a$}}
\put(76.33,25.67){\makebox(0,0)[cc]{$a^+/a\ra a$}}
%\sigma_8
\put(63.00,16.67){\makebox(0,0)[cc]{$8$}}
\put(73,12.67){\makebox(0,0)[cc]{$(a^2)^+/a^2\ra a$}}
\end{picture}
\caption{Routing constructs (from left to right): sequential, split, and join. }\label{routing-const-fig}
\end{figure}

%In this work we consider 
SNP systems where each neuron has exactly one rule are called \textit{simple}, while the systems that have the same set of rules are called \textit{homogeneous} \cite{homogen-snp}. In this work, if SNP systems only have rules of the restricted form $(a^k)^+/a^k \rightarrow a$ where $k$ is a non-negative integer, we refer to them as \textit{semi-homogeneous}. We only consider SNP systems $\Pi$ and $ \overline{\Pi}$ that are simple and semi-homogeneous, where their initial configurations have one spike in the initial neuron only, and no spike in every other neuron (as in Figure \ref{example_snp-fig}) in this work. We make no restrictions on the values of the delays in a (rule of a) neuron. %Furthermore, we call a neuron a \textit{source} neuron if it has no incoming synapses. (BUT no such neuron in ITERATION routing...)
The objective is to \textit{route} or move the single spike in the initial neuron through the system, towards the output neuron, and eventually to the environment. Spikes are routed via \textit{paths}, where a path consists of at least two neurons $\sigma_i, \sigma_j$ such that $(i,j) \in syn$. Using paths, we can have four basic routing constructs (referring to Figure \ref{routing-const-fig}): 
\begin{enumerate}
\item \textit{sequential} where, given at least two neurons $\sigma_1$, $\sigma_{2}$ such that $\sigma_{2}$ spikes only after $\sigma_1$ spikes and there is a path from $\sigma_1$ to $\sigma_{2}$, 
\item \textit{iteration}, where at least two neurons spike multiple (possibly an infinite) number of times and a loop is formed e.g. adding a synapse $(2,1)$ which creates a loop between $\sigma_1$ and $\sigma_2$, 
\item \textit{split}, where a spike from $\sigma_3$ is sent to at least two output neurons $\sigma_4$ and $\sigma_5$ and $(3,4),(3,5) \in syn$, 
\item \textit{join}, where spikes from at least two input neurons $\sigma_6, \sigma_7$ are sent to a neuron $\sigma_8$, where $(6, 8), (7, 8) \in syn$, so that $\sigma_8$ produces a spike only after accumulating spikes from $\sigma_7$ and $\sigma_8$. 
\end{enumerate}
Notice that iteration routing can be formed by combining the three other constructs.
%SNP systems considered in this work are those with neurons having exactly one rule only. 
%Our motivation in considering $C_0$'s where the initial neuron has only one spike is as follows: 
Also notice that if there exists a sequential path from $\sigma_i$ (with delay $d_1$) to $\sigma_j$ (with delay $d_2$) so that $d_1 < d_2$ and the number of spikes of the initial neuron $\sigma_1$ in $C_0$ is $n_1 > 1$, it is possible for some spikes to be lost. The reason is that it is possible for $\sigma_j$ to still be closed when spikes from $\sigma_i$ arrive. We avoid lost spikes by considering SNP systems where the initial neuron has only one spike. % $d_1 \geq d_2$ whenever a sequential path from $\sigma_i$ to $\sigma_j$ exists. %, and leave the cases where $d1 < d2$ as an open problem.
We say in this work that a $ \overline{\Pi}$  \textit{simulates} a $\Pi$ if two requirements are satisfied:
\begin{itemize}

\item[$ \mathbf{R_1':}$]  halting time of $\Pi$ spikes is the same halting time of $ \overline{\Pi}$,
\item[$ \mathbf{R_2':}$] number of spikes in the environment of $\Pi$ when $\Pi$ halts is equal to the number of spikes in the environment of $ \overline{\Pi}$ when $ \overline{\Pi}$ halts.
\end{itemize} 
  %We define the \textit{total runtime} of $\Pi$ and $ \overline{\Pi}$ as the total time needed for a spike to arrive to the sink neurons from the source neurons.

%%%%%%%%%%%%%%%%%%%%%%%%%%%%%%%%%%%%%%%%%%%%%%%%%%%%%%%%%%%%%%%%%%%
\section{Main results}\label{main-results-sect}

We begin presenting our results with a fundamental idea on sequential routing. %Unless otherwise stated, the SNP systems referred in our results are simple, semi-homogeneous SNP systems. Initial configurations are such that only the initial neuron has a spike. 

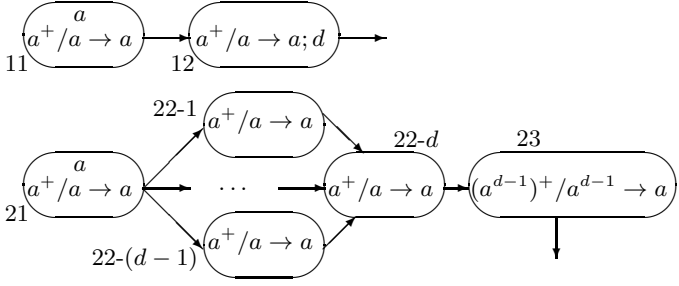
\begin{figure}[h]
\unitlength=1mm
\special{em:linewidth 0.4pt}
\linethickness{0.4pt}
\begin{picture}(116.33,40)(1,0)

%ovals i.e. neurons
\put(10.83,30.17){\oval(16.00,8.67)[]} %\sigma_11
\put(34.83,30.17){\oval(20.00,8.67)[]} %\sigma_12

\put(10.83,10.17){\oval(16.00,8.67)[]} %\sigma_21
\put(34.83,18.17){\oval(16.00,8.67)[]} %\sigma_22-1
\put(34.83,2.17){\oval(16.00,8.67)[]} %\sigma_22-(d-1)
\put(50.83,10.17){\oval(16.00,8.67)[]} %\sigma_23
\put(75.83,10.17){\oval(27.5,8.67)[]} %\sigma_24

%arcs i.e. synapses
\put(18.67,29.67){\vector(1,0){6.33}} %\sigma_11 -> \sigma_12
\put(44.67,29.67){\vector(1,0){6.33}} %12 -> env

\put(18.67,9.67){\vector(1,1){8}} %\sigma_21 -> 22-1
\put(18.67,9.67){\vector(1,0){6.33}} %\sigma_21 -> ...
\put(18.67,9.67){\vector(1,-1){8}} %\sigma_21 -> 22-(d-1)
\put(42.67,19.67){\vector(1,-1){5.33}} %\sigma_22-1 -> 22-d
\put(36.67,9.67){\vector(1,0){6.33}} %... -> 23
\put(42.67,1.67){\vector(1,1){4.33}} %\sigma_22-(d-1) -> 22-d
\put(58.67,9.67){\vector(1,0){3.33}} %22-d -> 23
\put(73.67,5.67){\vector(0,-1){5.33}} %23 -> env

%contents of neurons
%\sigma_11
\put(2,26.33){\makebox(0,0)[cc]{$11$}}
\put(10.33,32.67){\makebox(0,0)[cc]{$a$}}
\put(10.33,29.67){\makebox(0,0)[cc]{$a^+/a\ra a$}}
%\sigma_12
\put(24,26.33){\makebox(0,0)[cc]{$12$}}
\put(34.33,29.67){\makebox(0,0)[cc]{$a^+/a\ra a;d$}}

%\sigma_21
\put(2,6.33){\makebox(0,0)[cc]{$21$}}
\put(10.33,12.67){\makebox(0,0)[cc]{$a$}}
\put(10.33,9.67){\makebox(0,0)[cc]{$a^+/a\ra a$}}
%\sigma_22-1
\put(23,20.33){\makebox(0,0)[cc]{$22$-$1$}}
\put(34.33,18.67){\makebox(0,0)[cc]{$a^+/a\ra a$}}
% ...
\put(30.67,9.67){\makebox(0,0)[cc]{$ \ldots$}}
%\sigma_22-(d-1)
\put(19,0.33){\makebox(0,0)[cc]{$22$-$(d-1)$}}
\put(34.33,2.67){\makebox(0,0)[cc]{$a^+/a\ra a$}}
%\sigma_22-d
\put(55,16.33){\makebox(0,0)[cc]{$22$-$d$}}
\put(50.33,9.67){\makebox(0,0)[cc]{$a^+/a\ra a$}}
%\sigma_23
\put(70,16.33){\makebox(0,0)[cc]{$23$}}
\put(75.33,9.67){\makebox(0,0)[cc]{$(a^{d -1 })^+/a^{d -1 }\ra a$}}
\end{picture}
\caption{Sequential routing: $\Pi_1$ (top) with delay $d$, and $ \overline{\Pi}_1$ (bottom) simulating $\Pi_1$. }\label{sequential-fig}
\end{figure}

\begin{mylem}[Sequential routing]\label{sequential-lemma}
Given an SNP system with delay $\Pi$ performing sequential routing, there exists an SNP system without delay $ \overline{\Pi}$ peforming sequential routing that simulates $\Pi$.
\end{mylem}
\proof We refer to Figure \ref{sequential-fig} for illustrations. Let 

$\Pi = (O, \sigma_{11}, \sigma_{12}, \{ (11,12) \}, 12 )$ with 
$\sigma_1 = ( 1, a^+/a \rightarrow a )$ and $ \sigma_2 = (0, a^+/a \rightarrow a;d )$ 

we then let 

$ \overline{\Pi} = ( O, \sigma_{21}, \sigma_{22 \text{-}i}, \sigma_{23}, syn, 23 ) $ where $1 \leq i \leq d$, 
$ syn = \{ (21, 22 \text{-} 1)$, $\ldots, (21, 22 \text{-}(d-1))$, $(22 \text{-}1, 22 \text{-}d)$, $\ldots, (22 \text{-}(d-1),22 \text{-}d)$, $(22 \text{-} d, 23) \}$, 
$\sigma_{21} = ( 1, a^+/a \rightarrow a)$, $\sigma_{22 \text{-} i} = (0, a^+/a \rightarrow a)$,
 $\sigma_{23} = (0, (a^{d -1 })^+/a^{d -1} \rightarrow a).$ 
 
The additional $d$ neurons, immediately after initial neuron $\sigma_{21}$ in $ \overline{\Pi}$, are used to multiply the single spike from $\sigma_{21}$. The additional neurons then send one spike each to $\sigma_{22\text{-}d}$. Neuron $\sigma_{22 \text{-}d}$ accumulates $d-1$ spikes, and consumes these, one spike at a time and sending one spike every time to $ \sigma_{23}$. This consumption of one spike every time step creates a delay of $d-1$ time steps. Due to the regular expression of the rule in $\sigma_{23}$, the neuron will have to accumulate $d - 1$ spikes before the rule is used. Once $\sigma_{23}$ accumulates $d -1 $ spikes, it immediately sends one spike to the environment. This spiking and halting occurs at time $t+d+1$ for both $\Pi$ and $ \overline{\Pi}$ (satisfying $ \mathbf{R_1'}$ and $ \mathbf{R_2'}$) if we  let $t$ be the time when $\sigma_{11}$ and $\sigma_{21}$ spike. 

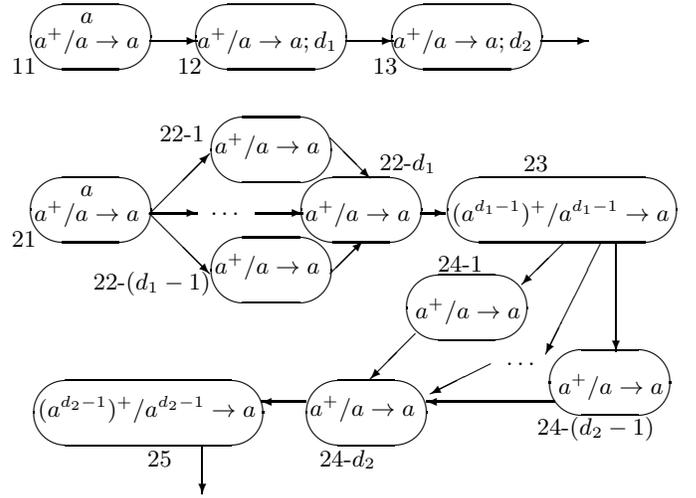
\begin{figure}[h]
\unitlength=1mm
\special{em:linewidth 0.4pt}
\linethickness{0.4pt}
\begin{picture}(116.33,65)(1,0)

%ovals i.e. neurons
\put(10.83,60.17){\oval(16.00,8.67)[]} %\sigma_11
\put(34.83,60.17){\oval(20.00,8.67)[]} %\sigma_12
\put(60.83,60.17){\oval(20.00,8.67)[]} %\sigma_13

\put(10.83,37.17){\oval(16.00,8.67)[]} %\sigma_21
\put(34.83,45.17){\oval(16.00,8.67)[]} %\sigma_22-1
\put(34.83,29.17){\oval(16.00,8.67)[]} %\sigma_22-(d_1-1)
\put(46.83,37.17){\oval(16.00,8.67)[]} %\sigma_22-d_1
\put(73.5,37.17){\oval(30.5,8.67)[]} %\sigma_23

\put(60.83,24.17){\oval(16,8.67)[]} %\sigma_24-1
\put(79.83,14.17){\oval(16.00,8.67)[]} %\sigma_24-(d_2-1)

\put(47.5,10.17){\oval(16,8.67)[]} %\sigma_24-d_2
\put(18.5,10.17){\oval(30.5,8.67)[]} %\sigma_25

%arcs i.e. synapses
\put(18.67,59.67){\vector(1,0){6.33}} %\sigma_11 -> \sigma_12
\put(44.67,59.67){\vector(1,0){6.33}} %12 -> 13
\put(70.67,59.67){\vector(1,0){6.33}} %12 -> env

\put(18.67,36.67){\vector(1,1){8}} %\sigma_21 -> 22-1
\put(18.67,36.67){\vector(1,0){6.33}} %\sigma_21 -> ...
\put(18.67,36.67){\vector(1,-1){8}} %\sigma_21 -> 22-(d-1)
\put(42.67,46.67){\vector(1,-1){5.33}} %\sigma_22-1 -> 22-d_1
\put(32.67,36.67){\vector(1,0){6.33}} %... -> 22-d_1
\put(42.67,28.67){\vector(1,1){4.33}} %\sigma_22-(d-1) -> 22-d_1
\put(54.67,36.67){\vector(1,0){3.33}} %22-d_1 -> 23
\put(73.67,32.67){\vector(-1,-1){5.5}} %23 -> 24-1
\put(78.67,32.67){\vector(-1,-2){7.33}} %23 -> ...
\put(80.67,32.67){\vector(0,-1){14.33}} %23 -> 24-(d_2 - 1)
\put(54,20.67){\vector(-1,-1){6}} %24-1 -> 24-d_2
\put(64,16.67){\vector(-2,-1){8}} %... -> 24-d_2
\put(72.5,11.67){\vector(-1,0){17}} %24-(d_2-1) -> 24-d_2
\put(39.5,11.67){\vector(-1,0){6}} %24-d_2 -> 25
\put(25.67,5.67){\vector(0,-1){6.33}} %25 -> env

%contents of neurons
%\sigma_11
\put(2,56.33){\makebox(0,0)[cc]{$11$}}
\put(10.33,62.67){\makebox(0,0)[cc]{$a$}}
\put(10.33,59.67){\makebox(0,0)[cc]{$a^+/a\ra a$}}
%\sigma_12
\put(24,56.33){\makebox(0,0)[cc]{$12$}}
\put(34.33,59.67){\makebox(0,0)[cc]{$a^+/a\ra a;d_1$}}
%\sigma_12
\put(50,56.33){\makebox(0,0)[cc]{$13$}}
\put(60.33,59.67){\makebox(0,0)[cc]{$a^+/a\ra a;d_2$}}

%\sigma_21
\put(2,33.33){\makebox(0,0)[cc]{$21$}}
\put(10.33,39.67){\makebox(0,0)[cc]{$a$}}
\put(10.33,36.67){\makebox(0,0)[cc]{$a^+/a\ra a$}}
%\sigma_22-1
\put(23,47.33){\makebox(0,0)[cc]{$22$-$1$}}
\put(34.33,45.67){\makebox(0,0)[cc]{$a^+/a\ra a$}}
% ...
\put(28.67,36.67){\makebox(0,0)[cc]{$ \ldots$}}
%\sigma_22-(d_1-1)
\put(19,27.33){\makebox(0,0)[cc]{$22$-$(d_1-1)$}}
\put(34.33,29.67){\makebox(0,0)[cc]{$a^+/a\ra a$}}
%\sigma_22-d_1
\put(53,43.33){\makebox(0,0)[cc]{$22$-$d_1$}}
\put(46.33,36.67){\makebox(0,0)[cc]{$a^+/a\ra a$}}
%\sigma_23
\put(70,43.33){\makebox(0,0)[cc]{$23$}}
\put(73.33,36.67){\makebox(0,0)[cc]{$(a^{d_1 -1 })^+/a^{d_1 -1 }\ra a$}}
%\sigma_24-1
\put(60,30){\makebox(0,0)[cc]{$24$-$1$}}
\put(61,23.67){\makebox(0,0)[cc]{$a^+/a\ra a$}}
%\ldots
\put(68,16.67){\makebox(0,0)[cc]{$\ldots$}}
%\sigma_24-(d_2 - 1)
\put(78,8){\makebox(0,0)[cc]{$24$-$(d_2 - 1)$}}
\put(80,13.67){\makebox(0,0)[cc]{$a^+/a\ra a$}}
%\sigma_24-(d_2)
\put(45,4){\makebox(0,0)[cc]{$24$-$d_2$}}
\put(47,10.67){\makebox(0,0)[cc]{$a^+/a\ra a$}}
%\sigma_25
\put(20,4){\makebox(0,0)[cc]{$25$}}
\put(18.33,10.67){\makebox(0,0)[cc]{$(a^{d_2 -1 })^+/a^{d_2 -1 }\ra a$}}

\end{picture}
\caption{Sequential routing with multiple delays:  $\Pi_2$ (top) with delays $d_1$ and $d_2$,  and  $ \overline{\Pi}_2$ (bottom) simulating $\Pi_2$.}\label{sequential_multiple-fig}
\end{figure}

We can repeatedly apply the previous construction if there exist more than one neuron with a (rule having a) delay in a sequential path as seen in Figure \ref{sequential_multiple-fig}. 
It can be easily shown that if there exists $\sigma_i$ without delay in a sequential path between $\sigma_{11}$ and $\sigma_{12}$, the time to halt for both $\Pi_1$ and $ \overline{\Pi}_1$ still coincide. In particular, every additional $\sigma_i$ having a rule without a delay adds one time step to the halting time of both $\Pi$ and $ \overline{\Pi}$. Both $ \mathbf{R_1'}$ and $ \mathbf{R_2'}$ are still satisfied. \qed

A sample computation of $\Pi_1$ and $ \overline{\Pi}_1$ is shown in Table \ref{sequential-table}. For sample computations of $\Pi_2$ and $ \overline{\Pi}_2$ we refer to Table \ref{sequential-2delays-table}. From Lemma \ref{sequential-lemma} we have the following observation.

\begin{myobserve}\label{additional-neurons-seq-observe}
If $\Pi$ has more than one neuron with a delay in a rule, the total additional neurons in $ \overline{\Pi}$ is $ \sum \limits_{i =1 }^m d_i$ where $d_i$ is the delay of the rule in $\sigma_i$.
\end{myobserve}
%prove this next time as a corollary...?

\begin{table}[h]
	\centering
	\begin{tabular}{||c|c|c||}
	
	\hline 
	
	Steps & $\Pi_1$ & $\overline\Pi_1$ \\ \hline
	
	$t_0$ & $ \langle 1/0, 0/0, 0 \rangle $ & $ \langle 1, 0, 0, 0, 0, 0 \rangle $ \\ \hline
	
	$t_1$ & $ \langle 0/0, 1/0, 0 \rangle $ & $ \langle 0, 1, 1, 0, 0, 0 \rangle $ \\ \hline
	
	$t_2$ & $ \langle 0/0, 0/3, 0 \rangle $ & $ \langle 0, 0, 0, 2, 0, 0 \rangle $ \\ \hline
	
	$t_3$ & $ \langle 0/0, 0/2, 0 \rangle $ & $ \langle 0, 0, 0, 0, 1, 0 \rangle $ \\ \hline
	
	$t_4$ & $ \langle 0/0, 0/1, 0 \rangle $ & $ \langle 0, 0, 0, 0, 2, 0 \rangle $ \\ \hline
	
	$t_5$ & $ \langle 0/0, 0/0, 1 \rangle $ & $ \langle 0, 0, 0, 0, 0, 1 \rangle $ \\ \hline
	
	\end{tabular}
	
	\caption{Sample computations of $\Pi_1$ and $ \overline{\Pi}_1$, $d = 3$.}\label{sequential-table}
\end{table}

\begin{table}[h]
	\centering
	\begin{tabular}{||c|c|c||}
	\hline
	
	Steps & $\Pi_2$ & $\overline\Pi_2$ \\ \hline
	
	$t_0$ & $ \langle $1/0, 0/0, 0/0, 0$ \rangle $ & $ \langle $1, 0, 0, 0, 0, 0, 0, 0, 0$ \rangle $ \\ \hline
	
	$t_1$ & $ \langle $0/0, 1/0, 0/0, 0$ \rangle $ & $ \langle $0, 1, 0, 0, 0, 0, 0, 0, 0$ \rangle $ \\ \hline
	
	$t_2$ & $ \langle $0/0, 0/2, 0/0, 0$ \rangle $ & $ \langle $0, 0, 1, 0, 0, 0, 0, 0, 0$ \rangle $ \\ \hline
	
	$t_3$ & $ \langle $0/0, 0/1, 0/0, 0$ \rangle $ & $ \langle $0, 0, 0, 1, 0, 0, 0, 0, 0$ \rangle $ \\ \hline
	
	$t_4$ & $ \langle $0/0, 0/0, 1/0, 0$ \rangle $ & $ \langle $0, 0, 0, 0, 1, 1, 0, 0, 0$ \rangle $ \\ \hline
	
	$t_5$ & $ \langle $0/0, 0/0, 0/3, 0$ \rangle $ & $ \langle $0, 0, 0, 0, 0, 0, 2, 0, 0$ \rangle $ \\ \hline
	
	$t_6$ & $ \langle $0/0, 0/0, 0/2, 0$ \rangle $ & $ \langle $0, 0, 0, 0, 0, 0, 1, 1, 0$ \rangle $ \\ \hline
	
	$t_7$ & $ \langle $0/0, 0/0, 0/1, 0$ \rangle $ & $ \langle $0, 0, 0, 0, 0, 0, 0, 2, 0$ \rangle $ \\ \hline
	
	$t_8$ & $ \langle $0/0, 0/0, 0/0, 1$ \rangle $ & $ \langle $0, 0, 0, 0, 0, 0, 0, 0, 1$ \rangle $ \\ \hline
	
	\end{tabular}
	\caption{Sample computations of $\Pi_2$ and $ \overline{\Pi}_2$, $d_1 = 2$, $d_2 = 3$.}
	\label{sequential-2delays-table}
\end{table}

\begin{figure}[h]
\unitlength=1mm
\special{em:linewidth 0.4pt}
\linethickness{0.4pt}
\begin{picture}(116.33,50)(1,0)

%ovals i.e. neurons
\put(18.83,45.17){\oval(16.00,8.67)[]} %\sigma_11
\put(42.83,45.17){\oval(20.00,8.67)[]} %\sigma_12

\put(11.83,20.17){\oval(16.00,8.67)[]} %\sigma_21-1
\put(36.83,20.17){\oval(16.00,8.67)[]} %\sigma_22-1
\put(36.83,2.17){\oval(16.00,8.67)[]} %\sigma_22-(d-1)
\put(61.83,20.17){\oval(16.00,8.67)[]} %\sigma_22-d
\put(36.83,32.17){\oval(30.00,8.67)[]} %\sigma_23

%arcs i.e. synapses
\put(26.67,43.67){\vector(1,0){6.33}} %\sigma_11 -> \sigma_12
\put(33.00,45.67){\vector(-1,0){6.33}} %\sigma_12 -> \sigma_11
\put(52.67,44.67){\vector(1,0){6.33}} %12 -> env

\put(19.67,20.67){\vector(1,0){9.33}} %21 -> 22-1
\put(19.67,20.67){\vector(1,-1){11}} %21 -> ...
\put(19.67,20.67){\vector(1,-2){9}} %21 -> 22-(d-1)
\put(44.67,20.67){\vector(1,0){9.33}} %22-1 -> 22-d
\put(42.67,8.67){\vector(1,1){11.33}} %... -> 22-d
\put(45,2.67){\vector(1,1){13}} %21-(d-1) -> 22-d
\put(55.5,23.67){\vector(-1,1){5}} %22-d -> 23 
\put(22,30.67){\vector(-1,-1){6}} %23 -> 21
\put(51.5,30.67){\vector(1,0){6}} %23 -> env

%contents of neurons
%\sigma_11
\put(10,41.33){\makebox(0,0)[cc]{$11$}}
\put(18.33,47.67){\makebox(0,0)[cc]{$a$}}
\put(18.33,44.67){\makebox(0,0)[cc]{$a^+/a\ra a$}}
%\sigma_12
\put(32,41.33){\makebox(0,0)[cc]{$12$}}
\put(43.33,44.67){\makebox(0,0)[cc]{$a^+/a\ra a;d$}}
%\sigma_21
\put(11,14.33){\makebox(0,0)[cc]{$21$}}
\put(12.33,21.67){\makebox(0,0)[cc]{$a$}}
\put(12,18.67){\makebox(0,0)[cc]{$a^+/a\ra a$}}
%\sigma_22-1
\put(35,14.33){\makebox(0,0)[cc]{$22$-$1$}}
\put(37,18.67){\makebox(0,0)[cc]{$a^+/a\ra a$}}
%...
\put(36,9.33){\makebox(0,0)[cc]{$\ldots$}}
%\sigma_22-(d-1)
\put(21,1.33){\makebox(0,0)[cc]{$22$-$(d-1)$}}
\put(37,2.67){\makebox(0,0)[cc]{$a^+/a\ra a$}}
%\sigma_22-d
\put(61,14.33){\makebox(0,0)[cc]{$22$-$d$}}
\put(62,18.67){\makebox(0,0)[cc]{$a^+/a\ra a$}}
%\sigma_23
\put(53,35.33){\makebox(0,0)[cc]{$23$}}
\put(37,30.67){\makebox(0,0)[cc]{$(a^{d-1})^+/a^{d-1}\ra a$}}

\end{picture}
\caption{Iteration routing: $\Pi_3$ (top) has a delay $d$ simulated by $ \overline{\Pi}_3$ (bottom). }\label{iteration-fig}
\end{figure}
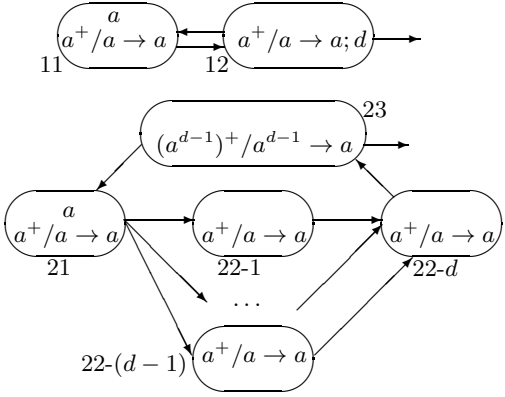

\begin{mylem}[Iteration routing]\label{iteration-lemma}
Given an SNP system with delay $\Pi$ performing iteration routing, there exists an SNP system without delay $ \overline{\Pi}$ peforming iteration routing that simulates $\Pi$.
\end{mylem}
\proof We refer to Figure \ref{iteration-fig} for illustrations. Let 

$\Pi = ( O, \sigma_{11},$ $\sigma_{12}, \{ (11,12),$ $(12, 11)$ $\}, 12 ) $ where $ \sigma_{11}$ = $(1,$ $a^+/a \rightarrow a;d$, $ \sigma_2 = (0, a^+/a \rightarrow a)$

we then let

$ \overline{\Pi} = ( O, \sigma_{21}, \sigma_{22 \text{-}i},$  $\sigma_{23},$ $syn, 23 ) $ where $\sigma_{21}$ = $(1,$ $a^+/a \rightarrow a )$, $\sigma_{22 \text{-}i}  = (0, a^+/a \rightarrow a)$ for $ 1 \leq i \leq d$, $ \sigma_{23} = (0, (a^{d-1})^+/a^{d-1} \rightarrow a )$ and $ syn = \{ ( 21, 22 \text{-} 1),$ $\ldots,$ $(21, 22  \text{-} (d-1) ),$ $(22  \text{-} 1, 22  \text{-} d),$ $\ldots,$ $( 22  \text{-} (d-1), 22  \text{-} d),$ $(22  \text{-} d, 23),$ $(23, 21) \},$ $out = 23 $

The construction of $ \overline{\Pi}$ uses the construction idea in Lemma \ref{sequential-lemma} i.e. the neuron with a delay $d$ in $\Pi$ is replaced with $d$ additional neurons in $ \overline{\Pi}$. In Figure \ref{iteration-fig} an infinite loop is created: a spike starts at $\sigma_{11}$ and it uses its rule at time $t$ so that the spike is sent to $\sigma_{12}$ at time $t+d$, then $\sigma_{12}$ immediately sends a spike back to $\sigma_{11}$ (and the environment) at time $t+d+1$, and so on and so forth. Similarly, $\sigma_{21}$ sends a spike to neurons $\sigma_{22 \text{-} 1}$ to $\sigma_{22  \text{-} (d-1)}$  at time $t$. At time $t+1$, $\sigma_{22  \text{-} d}$ accumulates $d-1$ spikes from the $d-1$ neurons from the previous time step. The spikes in $\sigma_{22  \text{-} d}$ are consumed and then sent one at a time to $\sigma_{23}$. At time $t+d$, $\sigma_{23}$ accumulates $d-1$ spikes so that it sends one spike back to $\sigma_{21}$ and at the environment at time $t+d+1$, coinciding with the time of spiking of $\sigma_{12}$. Thus, $ \mathbf{R_1'}$ and $ \mathbf{R_2'}$ are satisfied.
\qed

Lemma \ref{iteration-lemma} for iteration routing makes use of the construction used in Lemma \ref{sequential-lemma} for sequential routing. This construction will again be used for the join and split routings as follows. From Lemma \ref{iteration-lemma} we have the following observation.

\begin{myobserve}\label{new-initial-neuron-observe}
If the initial neuron of $\Pi$ has a delay and its halting time is $t+d$, we add a new initial neuron $\sigma_{1'}$ in $\Pi$ with $(1',1) \in syn$ so that $\Pi$ halts at time $t+d+1$. We then add a new initial neuron similarly to $ \overline{\Pi}$ and modify its $syn$ (following Lemma \ref{sequential-lemma} construction) so that $ \overline{\Pi}$ halts at $t+d+1$ instead, simulating $\Pi$.
\end{myobserve}

Although the premise of Observation \ref{new-initial-neuron-observe} is different from our assumption in Section \ref{prelims-sect} that the initial neuron has no delay, the observation provides a solution on how to approach such a premise. For example, if $\sigma_{11}$ has a delay instead of $\sigma_{23}$ in $\Pi_3$, we add a new initial neuron to $\sigma_{11'}$ and a new synapse $ (11', 11) \in syn$ in $\Pi$. For $ \overline{\Pi}_3$, we modify it as follows: add a new initial neuron $\sigma_{21'}$ and $\sigma_{21}$ is replaced with $d$ neurons (instead of $\sigma_{22}$). The synapse set of $ \overline{\Pi}$ is changed to $ \{ (21', 21 \text{-} 1),$ $\ldots,$ $(21',  21 \text{-} (d-1)),$ $(21  \text{-} 1, 21  \text{-} d),$ $\ldots$, $(21  \text{-} (d-1), 21  \text{-} d),$ $( 21  \text{-} d, 22),$ $(22, 21  \text{-} 1),$ $\ldots,$ $(22, 21  \text{-} (d-1)) \}$, we remove $\sigma_{23}$ and have $\sigma_{22}$ as the output neuron instead. Both $\Pi_3$ and $ \overline{\Pi}_3$ halt at the same time at $t+d+2$.

\begin{figure}[h]
\unitlength=1mm
\special{em:linewidth 0.4pt}
\linethickness{0.4pt}
\begin{picture}(116.33,50)(1,0)

%ovals i.e. neurons
\put(15.83,45.17){\oval(20.00,8.67)[]} %\sigma_11
\put(15.83,35.17){\oval(20.00,8.67)[]} %\sigma_12
\put(43.5,40.17){\oval(25.00,8.67)[]} %\sigma_13

\put(12.83,25.17){\oval(13.00,6.67)[]} %\sigma_21
\put(33.83,25.17){\oval(18.00,4.67)[]} %\sigma_23-(1)
\put(31.83,15.17){\oval(17.00,4.67)[]} %\sigma_23-(d)

\put(59.83,15.17){\oval(30.00,6.67)[]} %\sigma_25

\put(12.83,5.17){\oval(13.00,6.67)[]} %\sigma_22
\put(33.83,5.17){\oval(18.00,4.67)[]} %\sigma_23-(d-1)

%arcs i.e. synapses
\put(26,45.67){\vector(1,-1){5}} %\sigma_11 -> 13
\put(25.5,33.67){\vector(1,1){5.5}} %\sigma_12 -> 13
\put(55.5,38.67){\vector(1,0){5.5}} %\sigma_13 -> env

\put(19,23.67){\vector(1,0){6.33}} %\sigma_21 -> \sigma_23-(1)
\put(15,21.67){\vector(0,-1){5.33}} %\sigma_21 -> ...
\put(18,22){\vector(1,-2){7.5}} %\sigma_21 -> 23-(d-1)
\put(35,22.67){\vector(0,-1){5.33}} %\sigma_23-1 -> \sigma_23-d
\put(19,3.67){\vector(1,0){6.33}} %\sigma_22 -> \sigma_23-(d-1)
\put(15,8.67){\vector(0,1){5.33}} %\sigma_22 -> ...
\put(18,8.3){\vector(1,2){7.33}} %\sigma_22 -> 23-1
\put(35,7.67){\vector(0,1){5}} %23-(d-1) -> 23-d
\put(16,15.33){\vector(1,0){7.5}} %... -> 23-d
\put(40.5,15.67){\vector(1,0){4.3}} %23-d -> 25
\put(75,15.67){\vector(1,0){6.3}} %25 -> env

%contents of neurons
%\sigma_11
\put(26,49.33){\makebox(0,0)[cc]{{$11$}}}
\put(14.33,47.67){\makebox(0,0)[cc]{{$a$}}}
\put(15.33,43.67){\makebox(0,0)[cc]{{$a^+/a\ra a$}}}
%\sigma_12
\put(26,31.33){\makebox(0,0)[cc]{{$12$}}}
\put(14.33,37.67){\makebox(0,0)[cc]{{$a$}}}
\put(15.33,33.67){\makebox(0,0)[cc]{{$a^+/a\ra a$}}}
%\sigma_13
\put(46,34.33){\makebox(0,0)[cc]{{$13$}}}
\put(43.33,39.67){\makebox(0,0)[cc]{{$(a^2)^+/a^2\ra a;d$}}}
%\sigma_22
\put(7,1.33){\makebox(0,0)[cc]{\tiny{$22$}}}
\put(12.33,7.67){\makebox(0,0)[cc]{\tiny{$a$}}}
\put(12.33,4.67){\makebox(0,0)[cc]{\tiny{$a^+/a\ra a$}}}
%\sigma_25
\put(54,10.33){\makebox(0,0)[cc]{\tiny{$24$}}}
\put(59.33,15.67){\makebox(0,0)[cc]{\tiny{$(a^{d-1})^+/a^{d-1}\ra a$}}}

%\sigma_21
\put(7,21.33){\makebox(0,0)[cc]{\tiny{$21$}}}
\put(12.33,27.67){\makebox(0,0)[cc]{\tiny{$a$}}}
\put(12.33,24.67){\makebox(0,0)[cc]{\tiny{$a^+/a\ra a$}}}
%\sigma_23-(1)
\put(28,21.33){\makebox(0,0)[cc]{\tiny{$23$-$1$}}}
\put(34.33,24.67){\makebox(0,0)[cc]{\tiny{$(a^2)^+/a^2\ra a^2$}}}
%...
\put(14,15.33){\makebox(0,0)[cc]{\tiny{$ \ldots$}}}
%\sigma_23-(d-1)
\put(28,1.33){\makebox(0,0)[cc]{\tiny{$23$-$(d-1)$}}}
\put(34.33,4.67){\makebox(0,0)[cc]{\tiny{$(a^2)^+/a^2\ra a^2$}}}
%\sigma_23-(d)
\put(30,11.33){\makebox(0,0)[cc]{\tiny{$23$-$d$}}}
\put(31.33,15.67){\makebox(0,0)[cc]{\tiny{$(a^2)^+/a^2\ra a$}}}

\end{picture}
\caption{Join routing : $\Pi_4$ (top) has delay $d$ simulated by $ \overline{\Pi}_4$ (bottom). }\label{join-ver2-fig}
\end{figure}
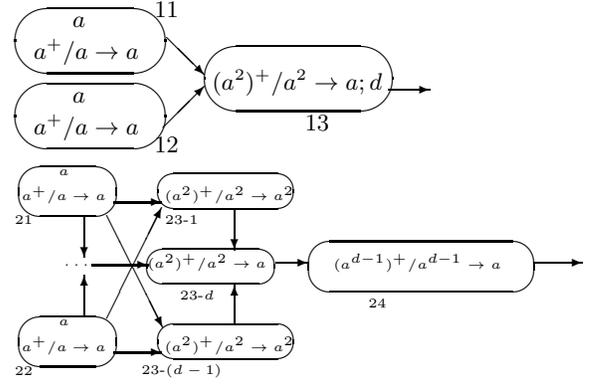

\begin{mylem}[Join routing]\label{join-lemma}
Given an SNP system with delay $\Pi$ performing join routing, there exists an SNP system without delay $ \overline{\Pi}$ peforming join routing that simulates $\Pi$.
\end{mylem}
\proof We refer to Figure \ref{join-ver2-fig} for illustrations. Let 

$\Pi = ( O, \sigma_{11}, \sigma_{12}, \sigma_{13}, \{ (11,13),(12,13)\}, 13  )$ where $\sigma_{11} = \sigma_{12} = (1, a^+/a \rightarrow a)$, $\sigma_{13} = (0, (a^2)^+/a^2 \rightarrow a;d)$

we then let

$ \overline{\Pi} = ( O, \sigma_{21}, \sigma_{22}, \sigma_{23 \text{-} i}, \sigma_{24}, syn, 24  ) $ where $ 1 \leq i \leq d$, $\sigma_{11} = \sigma_{22} = (1, a^+/a \rightarrow a)$, $ \sigma_{23 \text{-} 1} = \ldots = \sigma_{23 \text{-} (d-1)} = (0, (a^2)^+/a^2 \rightarrow a^2 )$, $ \sigma_{23 \text{-} d} = ( 0, (a^2)^+/a^2 \rightarrow a ) $, $syn = \{ (21, 23 \text{-}1),$ $\ldots,$ $(21, 23 \text{-} (d-1)),$ $(22, 23 \text{-} 1),$ $\ldots,$ $(22, 23 \text{-} (d-1)),$ $(23 \text{-} 1,$ $23 \text{-}d),$ $\ldots,$ $(23 \text{-} (d-1), 23 \text{-}d),$ $(23 \text{-} d, 24)  \}.$

%$ \overline{\Pi} = ( O, \sigma_{21}, \sigma_{22}, \sigma_{23 \text{-} i}, \sigma_{24 \text{-} i}, \sigma_{25}, syn, 25  ) $ where $1 \leq i \leq d$, $\sigma_{21} = \sigma_{22} = (1, a^+/a \rightarrow a)$, $ \sigma_{23 \text{-}i} = \sigma_{24 \text{-}i} = a^+/a \rightarrow a $, $ \sigma_{25} = (0, (a^{2(d-1)})^+/a^{2(d-1)} \rightarrow a;d $ and $ syn = \{ (21, 23\text{-}1),$ $\ldots,$ $(21, 23\text{-}(d-1)),$ $(23\text{-}1, 23\text{-}d),$ $\ldots,$ $(23\text{-}(d-1),$ $23\text{-}d),$ $(23\text{-}d, 25),$ $(22, 24\text{-}1),$ $(22, 24\text{-}(d-1)),$ $(24\text{-}1, 24\text{-}d),$ $(24\text{-}(d-1), 24\text{-}d),$ $(24\text{-}d, 25) \} $

For $\Pi$ and $ \overline{\Pi}$ we have as initial neurons $\sigma_{11}, \sigma_{12}$ and $\sigma_{21}, \sigma_{22}$ respectively. Using the construction in Lemma \ref{sequential-lemma}, $ \overline{\Pi}$ has $d$ additional neurons corresponding to $\sigma_{13}$ in $\Pi$. Let time $t$ be the time when the initial neurons spike. At time $t$, neurons $\sigma_{23 \text{-}1}$ to $\sigma_{23 \text{-}(d-1)}$ have two spikes each, so that in total, $ \overline{\Pi}$ at this time has $2(d -1)$ spikes. At the next time step $t+1$, the $d-1$ neurons send two spikes each to $\sigma_{23 \text{-} d}$ so that $\sigma_{23 \text{-} d}$ accumulates $2(d-1)$ spikes. Since $\sigma_{23 \text{-} d}$ consumes two spikes every time and it has $2(d-1)$ spikes, $\sigma_{23 \text{-} d}$ will take $d-1$ time steps to consume all of its $2(d-1)$ spikes. Every time $\sigma_{23 \text{-} d}$ spikes, it sends only one spike to $\sigma_{24}$. At time $t+d$, $\sigma_{24}$ has accumulated $d-1$ spikes from $\sigma_{23 \text{-} d}$ so that $\sigma_{24}$ sends one spike to the environment and halts at $t+d+1$. This time step coincides with the halting time of $\sigma_{13}$ in $\Pi$, sending one spike to the environment. We therefore satisfy $ \mathbf{R_1'}$ and $ \mathbf{R_2'}$. \qed

A sample computation of $\Pi_4$ and $ \overline{\Pi}_4$ is shown in Table \ref{join-ver2-table}. %Note that for iteration and join routing, Observation \ref{additional-neurons-seq-observe} still holds (For SPLIT????)....

\begin{table}[h]
	\centering
	\begin{tabular}{||c|c|c||}
	
	\hline 
	
	Steps & $\Pi_4$ & $\overline\Pi_4$ \\ \hline
	
	$t_0$ & $ \langle 1/0, 1/0, 0/0, 0 \rangle $ & $ \langle 1, 1, 0, 0, 0, 0, 0 \rangle $ \\ \hline
	
	$t_1$ & $ \langle 0/0, 0/0, 2/0, 0 \rangle $ & $ \langle 0, 0, 2, 2, 0, 0, 0 \rangle $ \\ \hline
	
	$t_2$ & $ \langle 0/0, 0/0, 0/3, 0 \rangle $ & $ \langle 0, 0, 0, 0, 4, 0, 0 \rangle $ \\ \hline
	
	$t_3$ & $ \langle 0/0, 0/0, 0/2, 0 \rangle $ & $ \langle 0, 0, 0, 0, 2, 1, 0 \rangle $ \\ \hline
	
	$t_4$ & $ \langle 0/0, 0/0, 0/1, 0 \rangle $ & $ \langle 0, 0, 0, 0, 0, 2, 0 \rangle $ \\ \hline
	
	$t_5$ & $ \langle 0/0, 0/0, 0/0, 1 \rangle $ & $ \langle 0, 0, 0, 0, 0, 0, 1 \rangle $ \\ \hline
	
	\end{tabular}
	
	\caption{Sample computations of $\Pi_4$ and $ \overline{\Pi}_4$, $d = 3$.}\label{join-ver2-table}
\end{table}

\begin{mylem}[Split routing]\label{split-lemma}
Given an SNP system with delay $\Pi$ performing split routing, there exists an SNP system without delay $ \overline{\Pi}$ peforming split routing that simulates $\Pi$.
\end{mylem}
\proof We refer back to the split routing in Figure \ref{routing-const-fig}. Notice that a split routing can be thought of as two separate paths, either from $\sigma_3$ to $\sigma_4$ or $\sigma_3$ to $\sigma_5$.  We let $t$ be the time that $\sigma_3$ spikes and modify the split routing in Figure \ref{routing-const-fig} as follows and let it be
 
$\Pi = ( O, \sigma_3, \sigma_4, \sigma_5, \sigma_o, syn, o )$ where $ \sigma_3 = (1, a^+/a \rightarrow a)$, $\sigma_5 = (0, a^+/a \rightarrow a)$, $ \sigma_4 = (0, a^+/a \rightarrow a;d)$, $syn = \{ (3,4),$ $(3,5),$ $(4,o),$ $(5,o)  \}$.

We arbitrarily chose $\sigma_4$ to have a delay instead of $\sigma_5$ in this case. Next we let

$ \overline{\Pi} = ( O, \sigma_{3'}, \sigma_{4' \text{-} i}, \sigma_{5'}, \sigma_{o}, syn, o )  $ where $1 \leq i \leq d$, $ \sigma_{3'} = (1, a^+/a \rightarrow a)$, $ \sigma_{4'  \text{-} 1} = \ldots = \sigma_{4'  \text{-}  (d-1) } = \sigma_{5'} = (0, a^+/a \rightarrow a)$, $ \sigma_{4'  \text{-} d} = (0, (a^{d-1})^+/ a^{d-1} \rightarrow a)$, $syn = \{ (3', 4'  \text{-} 1),$ $\ldots,$ $(3', 4'  \text{-} (d-1)),$ $(3', 5'),$ $(4'  \text{-} 1, 4'  \text{-} d),$ $\ldots,$ $(4'  \text{-} (d-1), 4'  \text{-} d),$ $(4'  \text{-} d, o),$ $(5,o)  \}$.

Since $\sigma_4$ has delay $d$, we simply follow Lemma \ref{sequential-lemma} and add $d$ neurons in $ \overline{\Pi}$ corresponding to $\sigma_4$. Let $t$ be the time when $\sigma_{3}$ and $ \sigma_{3'}$ spike. The time that $\sigma_o$ spikes the second time (since the spike from $\sigma_5$ makes $\sigma_o$ spike the first time, followed by the delayed spike from $\sigma_4$) i.e. the halting time, is $t+d + 1$ and the environment receives two spikes in total. This time coincides with the halting time of $ \overline{\Pi}$, which also sends two spikes to its environment. $ \mathbf{R_1'}$ and $ \mathbf{R_2'}$ are both satisfied for this case.

In the case where both $\sigma_4$ and $\sigma_5$ have delays $d_4$ and $d_5$ respectively, then following Lemma \ref{sequential-lemma}, $ \overline{\Pi}$ has $d_4 + d_5$ additional neurons. For both systems, halting time is $t+d_{max}+1$ where $d_{max} = max(d_4,d_5)$, and two spikes are sent to the environment, satisfying $ \mathbf{R_1'}$ and $ \mathbf{R_2'}$. \qed

We can now have the following theorem.

\begin{mythm}
Given an SNP system $\Pi$ with delays containing one or more of the following routings: sequential, iteration, join, split, there exists an SNP system $ \overline{\Pi}$ that simulates $\Pi$.
\end{mythm}
\proof Proof follows from Lemma \ref{sequential-lemma}, \ref{iteration-lemma}, \ref{join-lemma}, \ref{split-lemma}. \qed

Notice that Observation \ref{additional-neurons-seq-observe} and \ref{new-initial-neuron-observe} hold for all four routing constructs. 

%%%%%%%%%%%%%%%%%%%%%%%%%%%%%%%%%%%%%%%%%%%%%%%%%%%%%%%%%%%%%%%%%%%
\section{Final remarks}\label{final-remarks-sect}

We have presented an alternative construction of a $ \overline{\Pi}$ that simulates a given $\Pi$, improving the previous work so that we use only one initial spike for both systems. The halting time of both systems also exactly coincide with one another. The trade off is that there is an ``explosion'' of neurons in $ \overline{\Pi}$ for every delay in $\Pi$ i.e. we add $d_i$ neurons in $ \overline{\Pi}$ for every $\sigma_i$ that has a delay.

For our further work, we will consider nondeterministic SNP systems i.e. neurons having more than one applicable rule, since we only consider deterministic systems in this work. Minimization of the number of neurons of a $ \overline{\Pi}$ simulating a $\Pi$ is also desirable, including providing bounds to the number of neurons, spikes, and types of spiking rules. We will also use the matrix representation of SNP systems without delays from \cite{snp-mat} and then use massively parallel processors such as graphics processing units to create simulations of computations as was done in \cite{snpgpu-romjist}. Lastly, certain results and applications of SNP systems that use delays can be converted to SNP systems without delays e.g. generating automatic sequences as in \cite{snp-th-m}, and performing arithmetic operations as in \cite{arith-snp}, among others.

%%%%%%%%%%%%%%%%%%%%%%%%%%%%%%%%%%%%%%%%%%%%%%%%%%%%%%%%%%%%%%%%%%%
%\section*{Acknowledgments}

 %{Miguel A. Mart\'inez--del--Amor is supported by ``Proyecto de 

%%%%%%%%%%%%%%%%%%%%%%%%%%%%%%%%%%%%%%%%%%%%%%%%%%%%%%%%%%%%%%%%%%%


\begin{thebibliography}{4}\label{biblio}

\bibitem{snp-alhazov} Alhazov, A., Freund, R., Oswald, M., Slavkovik, M.: Extended Spiking Neural
P Systems. P\u aun, Gh., Rozenberg, G., Salomaa, A. (eds) Membrane computing, international workshop, WMC7,
revised, selected, and invited papers, Leiden, The Netherlands. LNCS, vol 4361. Springer, Berlin, pp 123-134, (2006)

%\bibitem{auto-seq} Allouche, J-P., Shallit, J. Automatic Sequences: Theory, Applications, Generalizations. Cambridge University Press (2003)

%\bibitem{snpgpu-cmc}  Cabarle, F.G.C.,  Adorna, H.N.,  Mart\'{i}nez-del-Amor, M.A.: A Spiking Neural P system simulator based on CUDA. M. Gheorghe et al. (eds.): 12th CMC 2011 revised and selected papers, LNCS 7184, pp. 87-103 (2012).

\bibitem{snpgpu-romjist}  Cabarle, F.G.C.,  Adorna, H.N.,  Mart\'{i}nez-del-Amor, M.A.,  P\'{e}rez-Jim\'{e}nez, M.J.: Improving GPU Simulations of Spiking Neural P Systems. Romanian Journal of Information Science and Technology, vol. 15(1) (2012)

\bibitem{snp-delays} Cabarle, F.G.C., Bu\~no, K.C., Adorna, H.N.: Time After Time: Notes on Delays In Spiking Neural P Systems. Workshop on Computation: Theory and Practice 2012 (WCTP 2012), 27-28 September, Manila, Philippines, preprint available online: {\tt http://arxiv.org/abs/1210.6119} (2012)

\bibitem{snp-th-m} Cabarle, F.G.C., Bu\~no, K.C., Adorna, H.N.: Spiking Neural P Systems Generating the Thue-Morse Sequence. Asian Conference on Membrane Computing, Wuhan, China, Oct (2012)

\bibitem{universality}  Chen, H., Ionescu, M., Ishdorj, T.-O.,  P\u aun, A., P\u aun, Gh.,  P\'{e}rez-Jim\'{e}nez, M.J.: Spiking neural P systems with extended rules: universality and languages. {Natural Computing}, vol. 7, issue 2, pp. 147-166 (2008)

%\bibitem{snp-cpu} Guti\' errez-Naranjo, Leporati, A.: First Steps Towards a CPU Made of Spiking Neural P Systems. Int. J. of Computers, Communications and Control. Vol. IV, No. 3, pp. 244-252 (2009)

%\bibitem{snp-ibarra-normal} Ibarra, O., P\u aun, A., P\u aun, Gh., Rodr\' iguez-Pat\' on, A., Sosik, P., Woodworth, S.  Normal forms for spiking neural P systems. Theory Comput Sci 372(2-3):196-217 (2007)


\bibitem{snp-ibarra} Ibarra, O.,  P\'{e}rez-Jim\'{e}nez, M.J., Yokomori, T.: On spiking neural P systems. Natural Computing, vol. 9, pp. 475-491 (2010)

\bibitem{snp} Ionescu, M., P\u aun, Gh., Yokomori, T.: Spiking Neural P Systems. {Fundamenta Informaticae}, vol. 71, issue 2,3 279-308, Feb. (2006)

\bibitem{compspikes} {Maass, W.}: Computing with spikes. { Special Issue on Foundations of
Information Processing of TELEMATIK}, vol 8(1), pp.32-36 (2002)

\bibitem{snpdivision} Pan, L.,  P\u aun, Gh.,  P\'{e}rez-Jim\'{e}nez, M.J.: Spiking neural P systems with neuron division and budding. {Proc. of the 7th Brainstorming  Week on Membrane Computing}, RGNC, Sevilla, Spain, pp. 151-168 (2009)

\bibitem{wsnp} Pan, L., Zeng, X., Zhang, X., Jiang, Y.: Spiking Neural P Systems with Weighted Synapses. Neural Processing Letters 35(1): 13-27 (2012)

\bibitem{introtomem} P\u aun, Gh.: {Membrane Computing: An Introduction}. Springer (2002)

\bibitem{snprecent} {P\u aun, Gh.,  P\'{e}rez-Jim\'{e}nez}, M.J.: Spiking Neural P Systems. Recent Results, Research Topics. A. Condon et al. (eds.), {Algorithmic Bioprocesses}, Springer (2009)

{
\bibitem{mem-handbook} {P\u aun, Gh., Rozenberg, G., Salomaa A. The Oxford Handbook of Membrane Computing, Oxford University Press (2010)}
}

{
\bibitem{homogen-snp} Zeng, X., Zhang, X., Pan, L.: Homogeneous Spiking Neural P Systems. Fundamenta Informaticae vol 97(1-2), pp. 275-294 (2009)
}

\bibitem{snp-mat} Zeng, X., Adorna, H.N., Mart\'{i}nez-del-Amor, M.A., Pan, L., P\'{e}rez-Jim\'{e}nez, M.J.: Matrix Representation of Spiking Neural P Systems. {11th CMC}, Jena, Germany, Aug. 2010 and {LNCS} 6501, pp. 377-39 (2011)

\bibitem{arith-snp} Zeng, X., Song, T., Zhang, X., Pan, L.: Performing Four Arithmetic Operations with Spiking Neural P Systems. IEEE Transaction on NanoBioscience. (in press)

%\bibitem{yao} A. Yao, \textit{Some complexity questions related to distributed computing}, Proceedings of the 11th ACM Symposium on Theory of Computing pp. 209-213 (1979).


\end{thebibliography}
\end{document}